\documentclass[letterpaper, 10 pt, conference]{ieeeconf}  
\usepackage{cite}

\IEEEoverridecommandlockouts                              

\usepackage{fancyhdr}
\fancypagestyle{withfooter}{
  
  \fancyfoot[C]{\footnotesize Accepted to the IEEE IROS workshop on Autonomous Robotic Systems in Aquaculture: Research Challenges and Industry Needs}
}

\usepackage{svg}
\usepackage{hyperref}
\usepackage{graphicx} 
\usepackage{tabularx}
\title{\LARGE \bf
Towards Design and Development of a Low-Cost Unmanned Surface Vehicle for Aquaculture Water Quality Monitoring in Shallow Water Environments*}

\author{Aiyelari Temilolorun$^{1}$ and Yogang Singh$^{2}$
\thanks{$^{1}$Aiyelari Temilolorun is with Department of Engineering and Mathematics, Sheffield Hallam University, Sheffield, S1 1WB, United Kingdom
        {\tt\small temi.aiyelari@gmail.com }}%
\thanks{$^{2}$Yogang Singh is with the Department of Engineering and Mathematics, Sheffield Hallam University, Sheffield, S1 1WB, United Kingdom
        {\tt\small y.singh@shu.ac.uk}}%
}

\begin{document}

\maketitle
\thispagestyle{withfooter}
\pagestyle{withfooter}

\begin{abstract}

 Unmanned surface vessels (USVs) are typically autonomous or remotely operated and are specifically designed for environmental monitoring in various aquatic environments. Aquaculture requires constant monitoring and management of water quality for the health and productivity of aquaculture systems. Poor water quality can lead to disease outbreaks, reduced growth rates, and even mass mortality of cultured species. Many small aquaculture operations, particularly in developing regions, operate on tight budgets and in shallow water environments such as inland ponds, coastal lagoons, estuaries, and shallow rivers. This leads to the foremost manoeuvrability challenge, underscoring the crucial need for agile cost-effective USVs as efficient monitoring systems. The paper proposes a low-cost 3D-printed twin-hull catamaran-style platform equipped with an Inertial Measurement Unit (IMU) and a Global Navigation Satellite System (GNSS) with a two-layered control framework and a differential drive configuration developed using two high-efficiency T-200 thrusters. The design utilizes the Robot Operating System (ROS) to create the control framework and incorporates Extended Kalman Filter (EKF)-based sensor fusion techniques for localisation. The paper evaluates the USV’s autonomy through open-water captive model experiments, employing remote control methods to assess the vessel’s manoeuvrability and overall performance characteristics in shallow water conditions.
\end{abstract}

\section{INTRODUCTION}

The global aquaculture industry has emerged as a pivotal contributor to food security, with its rapid growth significantly impacting the marine sector. In 2022, global aquaculture production soared to 185 million tons, generating an impressive 452 billion USD \cite{fisheries2024state}. Notably, developing regions in Asia, Africa, and South America have played a dominant role in this expansion, accounting for 89\% of the 80 million metric tons produced globally in 2016 \cite{tacon2018food}. This growth is largely driven by the proliferation of small-scale aquaculture operations, which are increasingly vital in meeting local demands \cite{garlock2020global}. However, these operations are often hindered by outdated practices, an ageing workforce, and labor shortages, which collectively degrade aquaculture water quality \cite{sarma83dam}.
Small-scale aquaculture in developing regions faces distinct challenges, particularly in shallow water environments. These environments are characterized by fluctuating temperatures, limited oxygen levels, and heightened susceptibility to contamination, all of which complicate water quality management \cite{ghose2014fisheries}. Traditional monitoring and intervention methods are frequently impractical in such settings, necessitating innovative approaches. Moreover, the high operational costs further exacerbate the difficulties faced by small-scale aquaculture in these regions \cite{naylor2023global}.
Addressing these challenges requires the development of cost-effective and agile robotic solutions. This manuscript introduces the design and development of a low-cost Unmanned Surface Vehicle (USV) tailored for water quality monitoring in shallow waters. These USVs are crucial for providing real-time data and management capabilities, ensuring the health and productivity of aquatic species while remaining accessible and affordable to small-scale operations in developing regions. The mobility of the USV, particularly in constrained shallow water environments, is essential for effective manoeuvrability where larger vessels struggle, making this an innovative and practical solution for the aquaculture industry.

The current manuscript explores an experimental investigation of proposed low-cost USV mobility using standard free-running maneuvering tests in shallow water, specifically focusing on the Turning Circle test \cite{eloot2015validation}. This test is crucial for assessing a USV's maneuverability in shallow water environments, as it measures the vehicle's ability to perform tight turns in restricted spaces. This is particularly important in areas where draft and clearance are limited. By analyzing the turning circle's size and the precision of the turns, the test provides valuable insights into how well the USV can navigate confined or shallow areas where larger vessels may face difficulties.

The remainder of the paper is structured as follows: Section \ref{Section1} covers the design and development of the USV platform, including hardware and control system design, cost analysis, software development, and an explanation of the EKF-based sensor fusion technique adopted in the current study. Section \ref{Section2} presents the experimental results for the designed USV platform, detailing the testing procedures, analysis from the free-running Turning Circle test, and key lessons learned. Finally, Section \ref{Section3} presents conclusions and future work.

\section{Design and Development of USV} \label{Section1}

\subsection{Hardware Design and Control System}

\begin{figure}[thpb]
      \centering
      \includegraphics[scale=0.5]{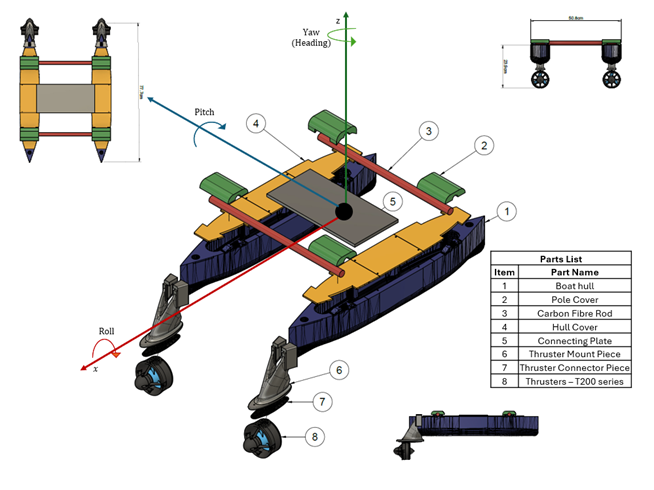}
      \caption{Assembled CAD model of the USV platform}
      \label{CADUSV}
   \end{figure}

The USV features a twin-hull catamaran design, 0.72m in length, and 0.41m in width. Each hull consists of three parts, 3D printed using Polylactic acid (PLA) filaments, spray-painted for waterproofing, and then joined together using Epoxy Resin to form a single unit. Carbon fibre rods are used to connect the two hulls. This lightweight construction allows for easy assembly of the boat. Figure \ref{CADUSV} illustrates the fully assembled computer-aided design (CAD) and a list of individual components.

\begin{figure}[thpb]
      \centering
      \includegraphics[scale=0.5]{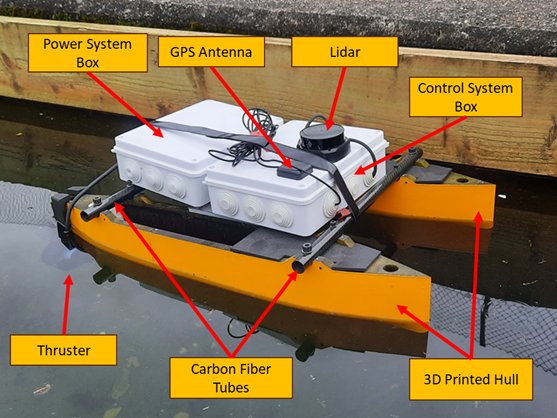}
      \caption{Developed prototype of the USV platform}
      \label{USVPrototype}
   \end{figure}

The assembled design is fitted with a power system, a sensors system and a propulsion system as shown in Figure \ref{USVPrototype} with the developed prototype. Electric power is supplied to the USV via 2 channels. The first channel is a power bank that provides the 5V required by the electronic system and sensors connected via USB to the Raspberry Pi (RPi) and the Arduino Mega. The second channel is the power for the propulsion system which is supplied by a Li-Po battery connected to outboard thrusters via a power distribution board and two electronic speed controllers (ESCs).

The propulsion system, depicted in Figure \ref{USVPrototype}, is powered by two high-efficiency T-200 thrusters. This design enables the vessel to be maneuvered with precision, independent of the fixed propellers’ reliance on rudders. Moreover, at very low speeds, rudders are inefficient for turning the vessel. The system employs a counter-rotating setup, with motors rotating in opposite directions, to balance forces, improve stability, and enhance maneuverability. This configuration enables the vessel to move straight, turn, or rotate in place by adjusting the relative speeds of the wheels.

\begin{figure}[thpb]
      \centering
      \includegraphics[scale=0.3]{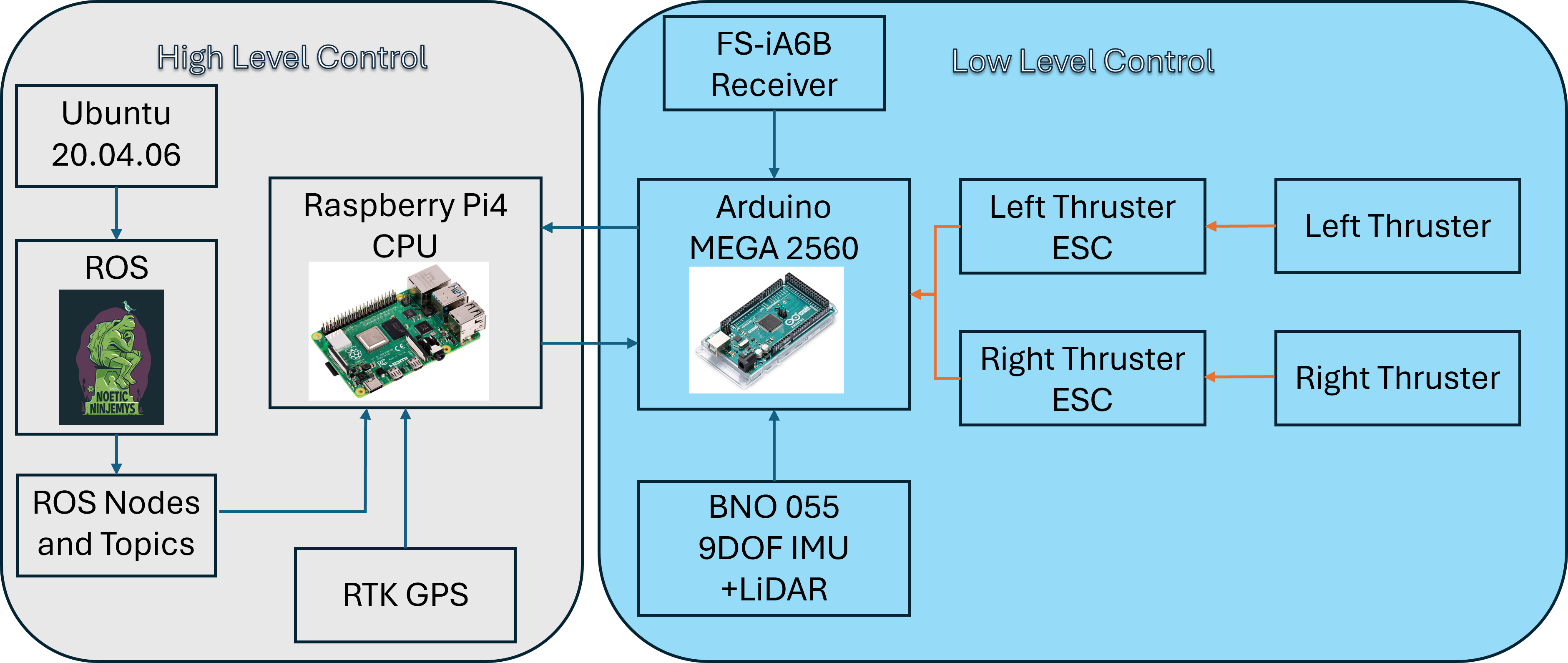}
      \caption{Schematic of the hardware design for the USV platform}
      \label{USVHardware}
   \end{figure}

The control system of the USV utilizes both high-level and low-level control mechanisms, achieved through the integration of a Raspberry Pi (RPi) 4 and an Arduino MEGA 2560, respectively. This dual-system approach allows for comprehensive control over the USV’s operations, encompassing both complex computational tasks and direct hardware interfacing. The low-level control system consists of an Arduino MEGA 2560 board as the main controller driving the ESCs, IMU and communication module. The thrusters are connected with specialized ESCs and controlled using pulse-width modulation on the Arduino MEGA. The ESCs are linked to a power distribution board (PDB), which in turn is connected to the Li-Po battery supplying power to the entire system. Figure \ref{USVHardware} illustrates the schematic of the hardware design for the USV platform.

The high-level control system consists of an RPi operating the ROS Noetic distribution within the Linux Ubuntu 20.04 environment. Through a serial connection, the GPS module interfaces directly with the Raspberry Pi, while the IMU sensor interfaces with the Arduino MEGA, and the combination of both establishes an \textit{ekf\textunderscore localisation} node within ROS. This node facilitates the localization of the boat within a three-dimensional space.
   
\subsection{Cost Analysis}

The design of our USV emphasizes a relatively low-cost construction while maintaining high functionality. Key electronic components, including a Raspberry Pi4, RTK GPS, IMU, Arduino Uno, Arduino MEGA, Lidar and RC Controller, collectively cost \$950 with the Lidar and RTK GPS costing more than half of the total amount. At the same time, the mechanical parts like IP7 electronics box, 3D Printing and Carbon Fibre tubes add \$160 to the total. This strategic selection of components results in a comprehensive, budget-friendly USV design with a cost of \$1118, making it an accessible option for research and development. The table \ref{Cost USV} summarises the cost breakdown for low-cost USV design.

\begin{table}[ht]
\caption{Cost breakdown for USV design}
\label{Cost USV}
\begin{center}
\begin{tabular}{c c c }
\hline
\textbf{Category} & \textbf{Items} & \textbf{Cost (\$)}\\
\hline
\textbf{Electronics} & T- 200 Thrusters (x2) & 200\\

& ESCs (x2) & 20\\
& RTK GPS & 350\\
&  IMU & 28 \\ 
&  Arduino Uno & 18 \\ 
& RC Controller &  60 \\ 
& 11.1V LiPo Battery & 24 \\ 
& Power Board &  8 \\ 
& Raspberry Pi4 & 50 \\ 
& Arduino MEGA & 50 \\ 
& Lidar & 150 \\ 

\textbf{Mechanical}& \small IP7 Electronics Box & \small 28   \\ 
& \small Carbon Fibre Tubes & \small 32   \\ 
& \small 3D Printing & \small 100 \\ 
\hline
& \textbf{Total} & 1118\\ 
\hline

\end{tabular}
\end{center}
\end{table}

\subsection{Software Design}

\begin{figure}[thpb]
      \centering
      \includegraphics[scale=0.32]{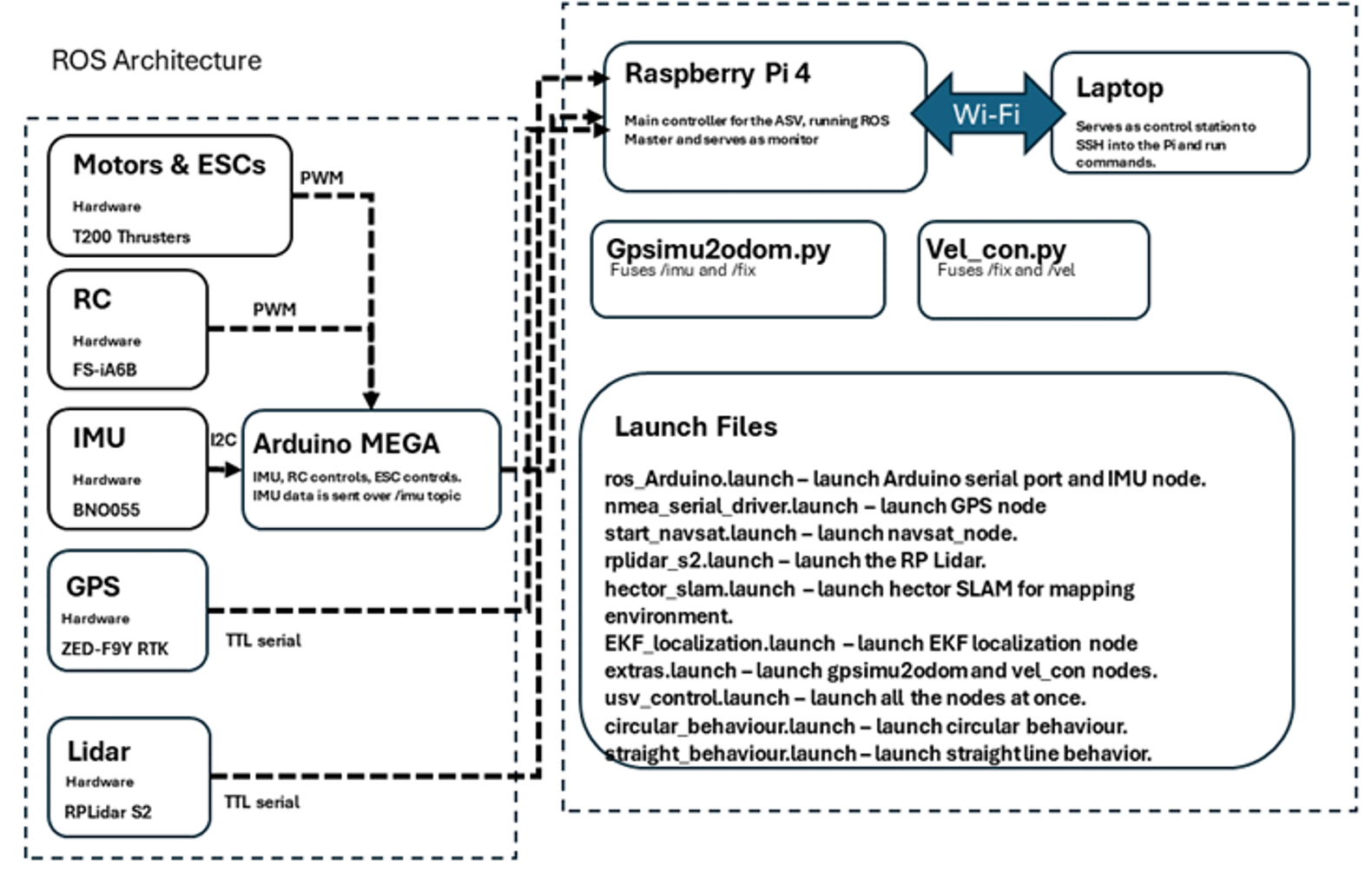}
      \caption{ROS architecture for the USV}
      \label{USVSoftware}
   \end{figure}

The software stack for the USV platform is built, utilizing a dual-layer approach that combines the strengths of both C++ and Python, in conjunction with the ROS as shown in Figure \ref{USVSoftware}. This structure ensures a robust and efficient software framework, catering to the distinct needs of both the low-level and high-level control systems of the USV.

The setup in ROS is built using nodes and launch files. Nodes are implemented to manage Arduino communication using ROSSERIAL and GPS communication via the serial driver. In addition, we have \textit{vel\textunderscore con} and \textit{gpsimu2odom} nodes that take data from the GPS and IMU publishers and transform that into odometry information. The RPi serves as the onboard computer and is responsible for running the ROS, and interfacing sensors. This setup is crucial for managing the autonomous control aspects of the USV. In ROS, the sensors connected are controlled using launch files to start them up while a subscriber is initiated to subscribe to the IMU data from the low-level controller. All this data is fed into the \textit{robot\textunderscore localization} package in the ROS which eventually contains the \textit{ekf\textunderscore localization} node to help the vessel better localize itself and perform state estimation. A detailed explanation of sensor fusion is explained in section \ref{EKF}.

\subsection{EKF based Sensor Fusion}\label{EKF}

In this study, we utilize the data fusion approach of \cite{moore2016generalized} to integrate data from an array of sensors on a USV. A \textit{robot\textunderscore localization} package developed as a part of the work has been used in the current study for the implementation of an EKF for sensor fusion. Although the original work was tested with mobile robots, this study extends its application to marine robots, adapting the methodology for use in the maritime domain. The current section details the implementation of the developed \textit{ekf\textunderscore localization} node as the component of the  \textit{robot\textunderscore localization} package. The goal of the EKF is to estimate the full 3D pose and velocity of the USV over time. The non-linear dynamical system of the USV can be described with 

\begin{equation}
    x_k = f(x_{k-1})+w_{k-1}
\end{equation}

where $x_k$ is the USV system state at a time $k$, $f$ is a non-linear state transition function and $w_{k-1}$ is the normally distributed noise. The USV state vector $x$ consists of the USV's 3D pose, orientation and their respective velocities. The received measurements are expressed as

\begin{equation}
    z_k = h(x_k)+v_k
\end{equation}

where $z_k$ is the measurement at time $k$, $h$ is a mapped non-linear sensor model and $v_k$ is the normally distributed measurement noise. In the initial stage, EKF performs a prediction step, advancing the current state estimate and error covariance forward in time using equations (3) and (4) as follows:

\begin{equation}
    \hat{x}_k = f(x_{k-1})    
\end{equation}

\begin{equation}
    \hat{P}_k = FP_{k-1} F^{T} + Q
\end{equation}

In this study, $f$ is the standard 3D kinematic model of the USV with $P$ being the estimated error covariance, projected via $F$, the Jacobian of $f$ and perturbed by $Q$, the process noise covariance. The correction step is then carried out using the equations (5) till (7) to calculate the Kalman gain using observation matrix $H$ and measurement covariance $R$ and $\hat{P}_k$ as follows:

\begin{equation}
    K = \hat{P}_k H^{T} (H\hat{P}_k H^{T} +R) ^{-1}
\end{equation}

\begin{equation}
    x_k = \hat{x}_k + K(z-H\hat{x}_k)
\end{equation}

\begin{equation}
   P_k = (I-KH)\hat{P}_k (I-KH)^{T}+KRK^{T}
\end{equation}

leading to the use of gain to update the state vector and covariance matrix. The fundamental aspect of the adopted approach is its capability for a partial update of the state vector, enabling the omission of certain variables in the state vector during the sensor data capture process. This capability is crucial for USVs, as data loss during operations is a common occurrence. In the present study, we consider the sensor configuration of fused odometry, incorporating a single IMU and a single GPS. This configuration is chosen to effectively manage the significant interference encountered in the operational environment of the USV.

\subsection{Communication System}\label{Communication}

The communication system of the USV is designed with two distinct parts namely, the Control Communications Part (CCP) and the Telemetry Communications Part (TCP) with both segments designed to employ radio communications to enhance the range and reliability of the data transmission. The CCP is primarily responsible for the remote-control operations of the USV. This system comprises two main elements: a transmitter and a receiver, both critical for the effective remote operation of the vehicle. The receiver in the CCP is intricately connected to the Arduino MEGA 2560, the core of the USV's control system and is mounted on the USV's electronics shelf. The placement of the receiver is crucial for optimal signal reception and processing, allowing for seamless communication between the remote operator and the USV.  The transmitter utilized in the CCP is the FS-i6 Flysky digital proportional radio control system. The choice of the FS-i6 Flysky transmitter is based on its efficiency, reliability, and compatibility with the overall design and operational requirements of the USV.

The TCP of the USV is strategically designed around the ROS communication between two computers, one on the robot and the local base station via a local wireless network, operating at 2.4GHZ ISM (Industrial, Scientific and Medical) band. This is to ensure minimal interference and efficient telemetry data transmission for both real-time analysis on the local base station and offline analysis through recorded sensor/ROS topics information saved on the robot using ROSBAGS.  With this setup, data transmitted via TCP between the two computers ensures reliable communication for the USV. The stable wireless connection between the two computers supports low-latency data exchange, which is crucial for real-time control and monitoring. Additionally, the recorded ROSBAGS enable detailed post-mission analysis, aiding in evaluating performance and diagnosing issues. This integrated system allows for precise and dependable operation of the USV, while also generating valuable data for ongoing improvements and system optimization.

\section{Experimental Results} \label{Section2}

\subsection{Testing Arena}
\begin{figure}[thpb]
      \centering
      \includegraphics[scale=0.41]{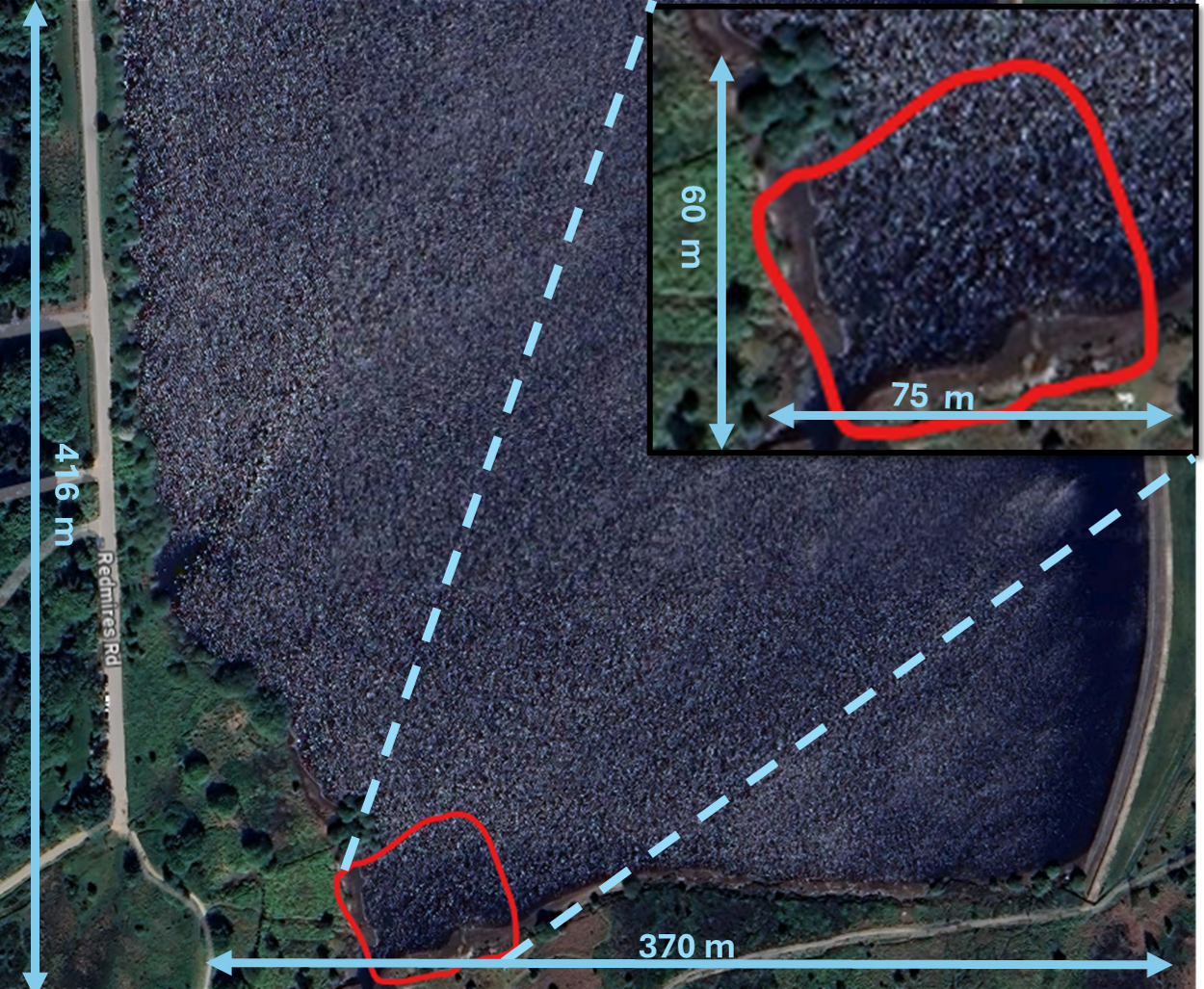}
      \caption{Marked testing arena for the USV platform}
      \label{TestingArena}
   \end{figure}

The USV platform was tested locally at Redmires Reservoir in Sheffield, United Kingdom. This location was selected for its realistic shallow water environment and proximity to local aquaculture operations, providing an ideal setting for practical testing. The specific testing area, highlighted in red in Figure \ref{TestingArena}, ensured that the USV could be evaluated in conditions that closely mimic its intended operational environment.

\subsection{Autonomy Protocol}

\begin{figure}[thpb]
      \centering
      \includegraphics[scale=0.37]{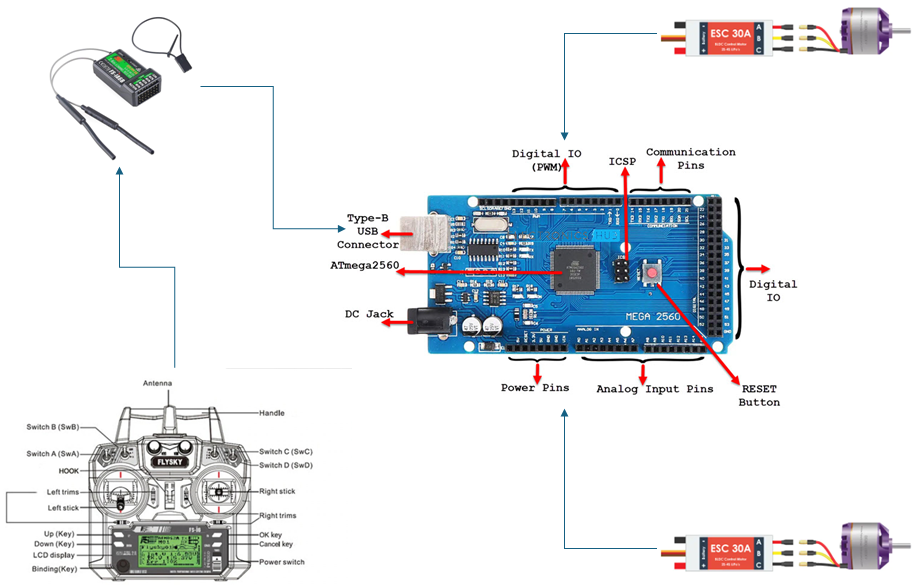}
      \caption{RC setup for the USV platform}
      \label{RCSet}
   \end{figure}

The vessel's autonomy was evaluated through a series of remote-control (RC) tests, utilizing radio communication to govern its motion as described in section \ref{Communication}. Sensor data recordings were made using ROSBAGs and analyzed using the \textit{MATLAB ROSBAG Viewer Tool} and \textit{Foxglove Studio}. The RC setup, illustrated in Figure \ref{RCSet}, enables the propulsion of the USV using a differential drive system. This system allows for precise movements, making the USV ideal for navigation in diverse aquatic environments. The tests were conducted using a laptop as a ground station with an Intel i7 1.90 GHz quad-core CPU and Ubuntu 20.04 as OS.

\subsection{Manoeuvring Experiments and Results}

\begin{figure}[thpb]
      \centering
      \includegraphics[scale=1.0]{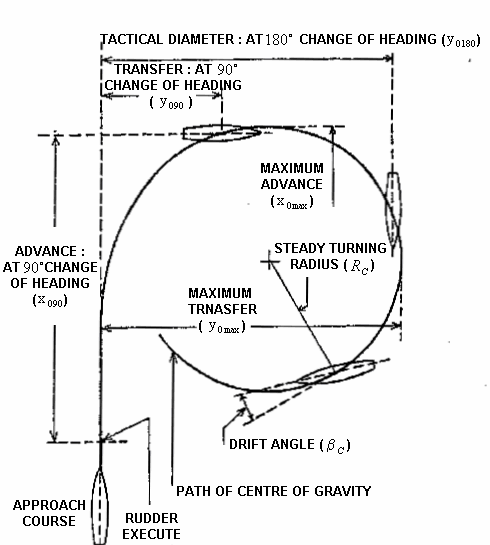}
      \caption{Turning Circle definitions (Port Side)}
      \label{Turn}
   \end{figure}

The manoeuvring performance of the vessel was evaluated in open water using a widely recognized captive model Turning Circle Test \cite{karimi2021guidance}. In the test, the vessel was made to do a turning circle of 540$^{\circ}$ to determine the following parameters as shown in Figure \ref{Turn}: 
\begin{itemize}
    \item Tactical diameter
    \item Advance
    \item Transfer
    \item Loss of speed on a steady turn
    \item Time to change heading 90$^{\circ}$
    \item Time to change heading 180$^{\circ}$
\end{itemize}
This test was conducted on the designed full-scale model of the USV within the testing arena depicted in Figure \ref{TestingArena}.

To perform the free-running manoeuvring experiments in open water, International Maritime Organisation (IMO) Resolution A.751 stating interim standards for ship maneuverability\cite{kim1996assessment, bakar2015manoeuvering} were adopted with suggested modifications for differential drive propulsion vessels as follows:

\begin{enumerate}
    \item The approach speed should be 90$\%$ of the vessel speed corresponding to 85$\%$ of the maximum engine output.
    \item Before manoeuvring, the vessel must maintain a steady course at a constant setting for at least one minute.
\end{enumerate}

\begin{figure}[thpb]
      \centering
      \includegraphics[scale=0.32]{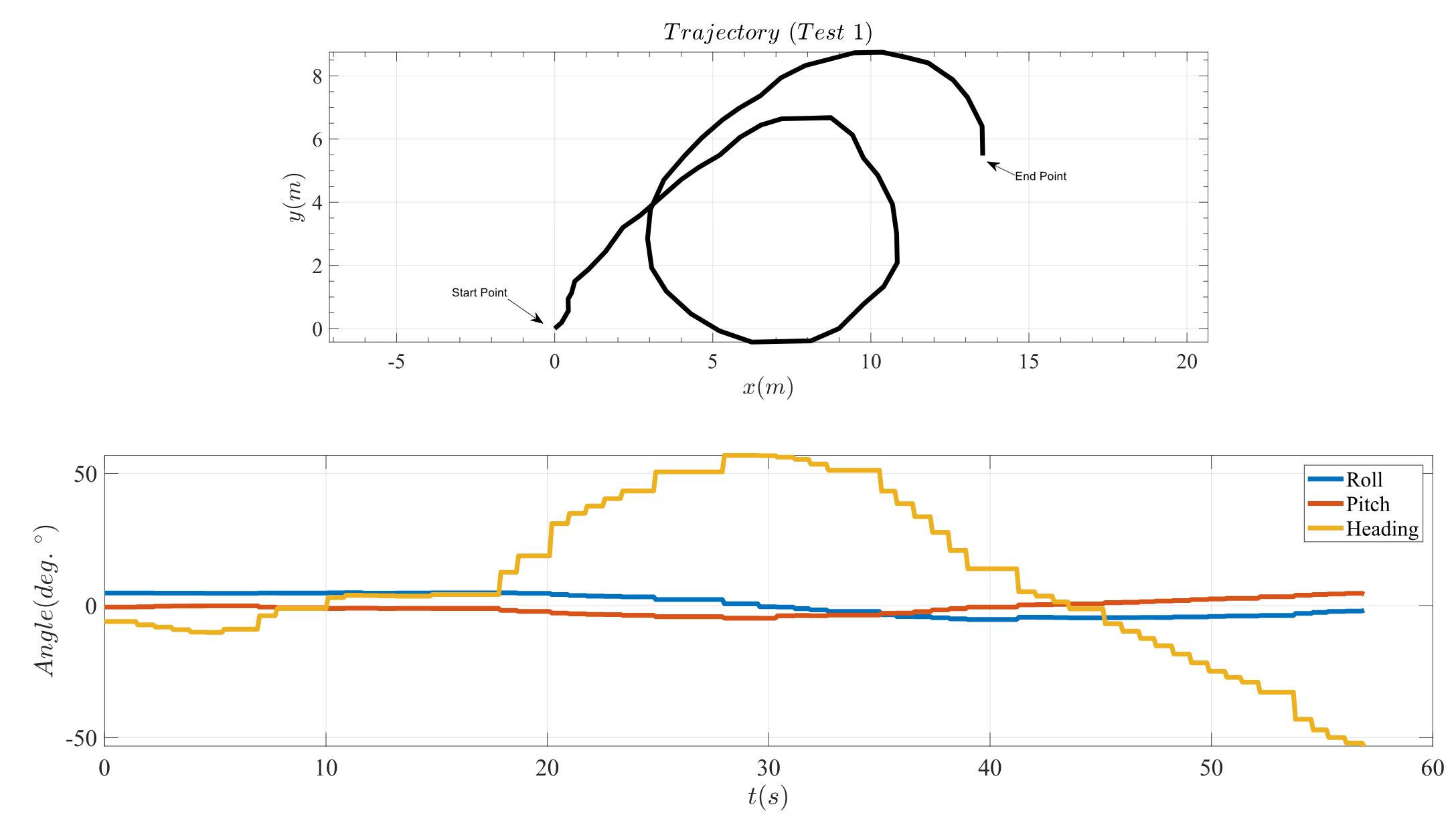}
      \caption{Turning Circle results from Test 1 (Starboard side)}
      \label{TurnS}
   \end{figure}

\begin{figure}[thpb]
      \centering
      \includegraphics[scale=0.32]{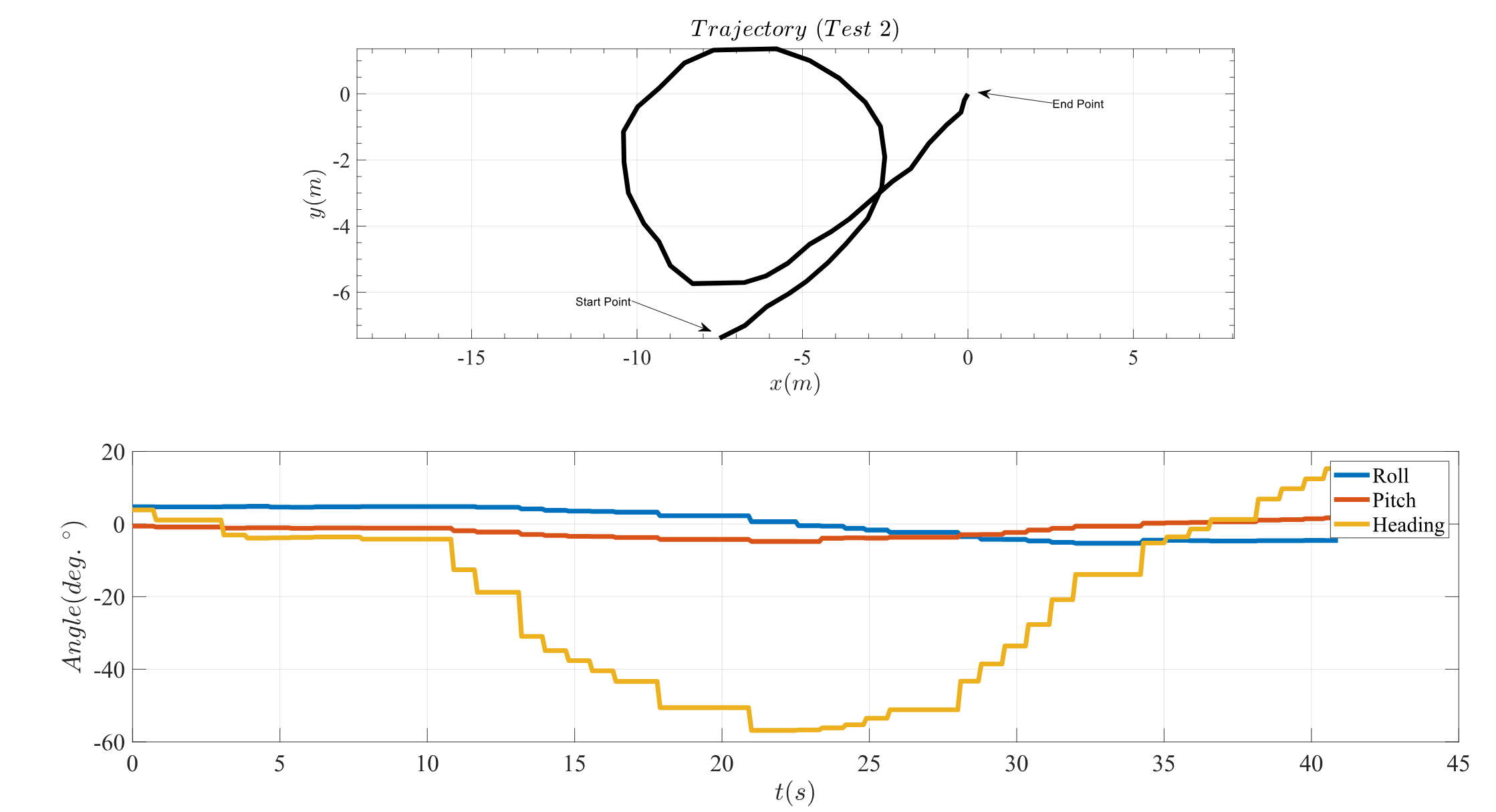}
      \caption{Turning Circle results from Test 2 (Port side)}
      \label{TurnP}
   \end{figure}

\begin{figure}[thpb]
      \centering
      \includegraphics[scale=0.3]{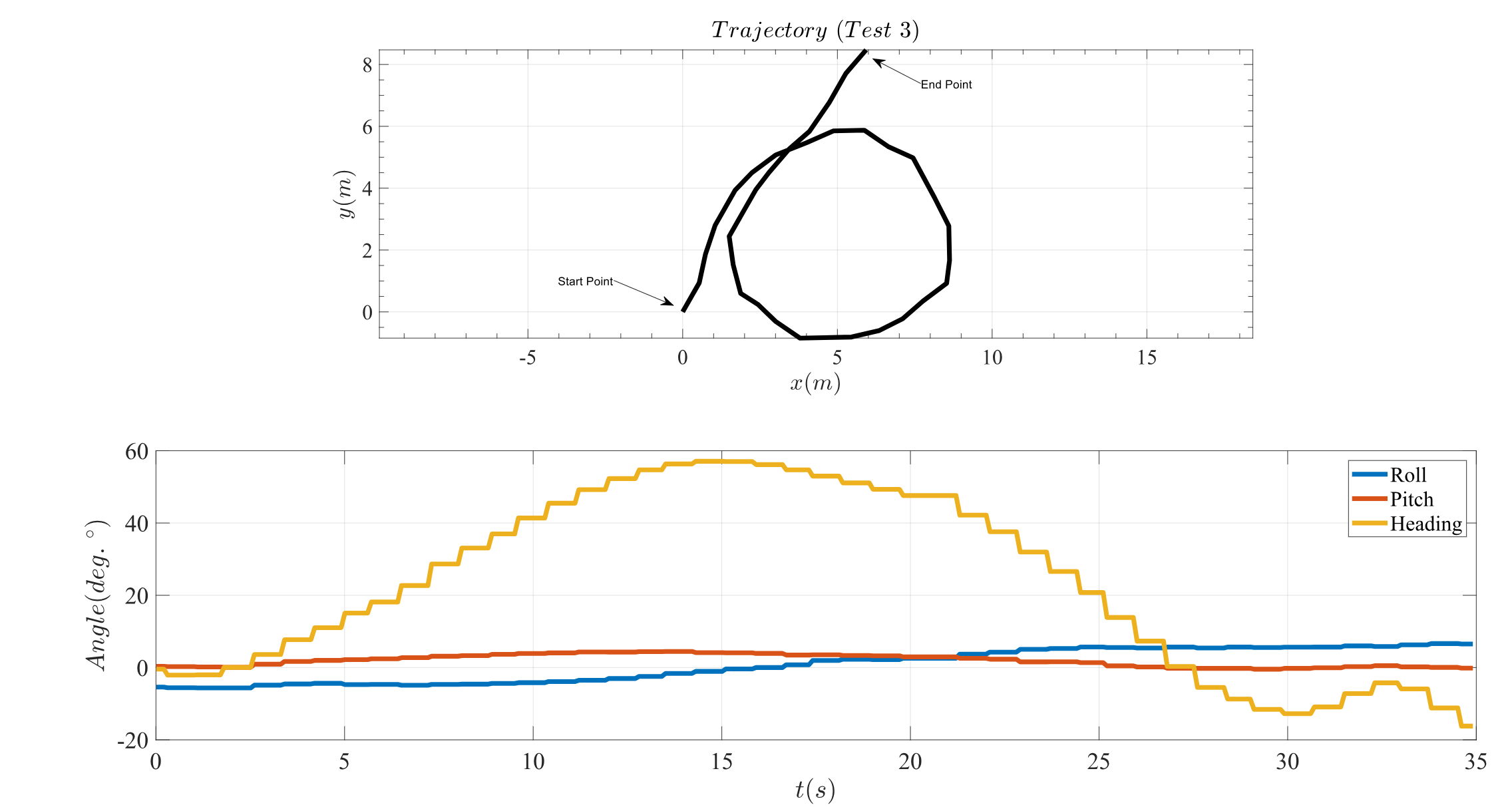}
      \caption{Turning Circle results from Test 3 (Starboard side)}
      \label{TurnS1}
   \end{figure}

Three different tests from three different start and goal points within the testing arena were performed to determine the USV manoeuvrability in terms of turning circle parameters and were compared against the IMO-enforced minimum manoeuvrability criteria listed in Table \ref{IMO} \cite{bakar2015manoeuvering}.
Figure \ref{TurnS} and Figure \ref{TurnP} present the data collected from the Turning Circle tests for the Starboard side while Figure \ref{TurnS1} presents the data collected on the Port side of the USV. The Turning Circle test data is summarised in Table \ref{Turning}. The IMO compliance for Tactical Diameter and Advance for the three tests is detailed in Table \ref{IMOTurning}. The trial data shows that the recorded values significantly exceed the criteria for meeting IMO standards regarding turning ability, particularly Tactical Diameter and Advance. This considerable deviation is attributed to the lower-cost sensor readings affecting measurement accuracy and external factors such as wind and surface currents in the testing area. Importantly, the measurements on the Starboard side display less variability compared to those on the Port side. The roll and pitch results from all three tests demonstrate minimal deviation, highlighting the stability and reliability of the USV, particularly in shallow waters and amidst challenging surface currents generated by wind. The recorded trial can be found at \href{https://www.youtube.com/watch?v=iVOl2lOuyIg}{https://www.youtube.com/watch?v=iVOl2lOuyIg} and the project repository can be found at \href{https://github.com/YogangSingh/catamaran_shu_v2}{https://github.com/YogangSingh/catamaran\textunderscore shu \textunderscore v2}.
\begin{table}[h]
\caption{IMO evaluation criteria for maneuverability \\(L- Length of the vessel)}
\label{IMO}
\begin{center}
\begin{tabular}{c c c }
\hline
\textbf{Parameters} & \textbf{Test} & \textbf{Criteria}\\
\hline
Advance & Turning Circle & $< $ 4.5L\\
Tactical Diameter & Turning Circle & $< $ 5L\\
\hline
\end{tabular}
\end{center}
\end{table}

\begin{table}[htbp]
\caption{Turning Circle test results \\ (P- Port side; S- Starboard side)}
\label{Turning}
\begin{center}
\begin{tabular}{c c c c}
\hline
\textbf{Parameters} & \textbf{Test 1 (S)} & \textbf{Test 2 (P)} & \textbf{Test 3 (S)}\\
\hline
Advance (m) & 7.03 & 6.10 & 5.88\\
Transfer (m) & 6.08 & 6.42 & 6.70\\
Tactical Diameter (m) & 5.16 & 4.99 & 6.67\\
Loss of speed-steady turn (\%) & 6.89 & 6.51 & 9.34\\
Time to change heading 90$^{\circ}$(s) & 11.52 & 12.27 & 8.74\\
Time to change heading 180$^{\circ}$(s) & 14.31 & 16.43 & 15.22 \\
\hline
\end{tabular}
\end{center}
\end{table}

\subsection{Lessons Learned}

The dual system approach using a Raspberry Pi and Arduino MEGA proved effective in managing both high level computational tasks and low-level hardware interfacing. The 3D printed hulls made from PLA filaments, reinforced with carbon fibre rods, demonstrated durability and stability in various water conditions, validating the feasibility of using cost-effective materials without compromising structural integrity. The Turning Circle tests highlighted significant deviations from the IMO criteria, indicating areas for improvement in sensor accuracy and propulsion control. The results showed that while the USV maintained good stability and minimal roll and pitch deviations, the tactical diameter and advance did not meet the required standards. The total cost of \$1118 for the USV, including electronics and mechanical parts, demonstrates that a functional USV can be developed on a low budget. Using widely available and affordable components like the Raspberry Pi, Arduino boards, and 3D-printed parts was pivotal in keeping the costs low while achieving the desired functionality. The USV achieved its aim of functional performance through successfully integrating navigation and control systems, even though the manoeuvrability metrics did not fully comply with IMO standards. 

\begin{table}[htbp]
\caption{IMO compliance for Turning Circle test \\ (P- Port side; S- Starboard side; TD- Tactical Diameter; A- Advance; Y- Yes; N- No)}
\label{IMOTurning}
\begin{center}
\begin{tabular}{c c c c c}
\hline
\textbf{Test} & \textbf{Parameters} & \textbf{Measured} & \textbf{Criteria} & \textbf{IMO Compliance(Y/N)}\\
\hline
1 (S) & TD, A & 7.07, 8.42 & 3.6, 3.24 & N, N\\
2 (P) & TD, A & 7.2, 9.08 & 3.6, 3.24 & N, N\\
3 (S) & TD, A & 6.8, 5.74 & 3.6, 3.24 & N, N\\
\hline
\end{tabular}
\end{center}
\end{table}

\section{CONCLUSIONS AND FUTURE WORK} \label{Section3}

The assessment of the performance and manoeuvrability of the designed USV was conducted through an open-water free-running turning circle test. The trial outcomes demonstrate a notable deviation from the specified IMO criteria, underscoring the imperative for further refinement in both hardware and software design for the USV. This necessitates the incorporation of more advanced sensors and the exploration of innovative design avenues. The iterative process seeks to elevate the USV's capabilities, ensuring compliance with regulatory standards while enhancing its efficiency and reliability across diverse aquaculture operations in shallow water. Despite this, the ongoing focus on integrating CAD-designed components, 3D printed parts, and a basic sensor suite, alongside limited testing, lays a foundational framework for future enhancements.

\addtolength{\textheight}{-12cm}   





\section*{ACKNOWLEDGMENT}

This research is supported by the Sheffield Hallam Department of Engineering \& Mathematics Research \& Innovation Funding for the project titled ``\textit{Towards Design and Development of an Autonomous Marine Vessel}". For the purpose of open access, the author has applied a Creative Commons Attribution (CC BY) licence to any author accepted manuscript version arising from this submission.



\bibliographystyle{IEEEtran}
\bibliography{Main}

\end{document}